\documentclass[11pt]{article}

\usepackage[preprint]{acl}
\usepackage{times}
\usepackage{latexsym}
\usepackage[T1]{fontenc}
\usepackage[utf8]{inputenc}
\usepackage{microtype}
\usepackage{inconsolata}
\usepackage{graphicx}

\usepackage{hyperref}
\usepackage{booktabs}
\usepackage{amssymb}
\usepackage{makecell}
\usepackage{pifont}
\usepackage{arydshln}
\usepackage{multirow}
\usepackage{amsmath}

\newcommand{\ours}{X-Dubbing}

\title{\ours{}: \\ Speech-to-Speech Translation under Temporal and Visual Constraints}

\author{Jeongsoo Choi$^1$ \quad Ji-Hoon Kim$^2$ \quad Junseok Ahn$^1$ \quad Joon Son Chung$^1$\\
$^1$KAIST \quad $^2$Chung-Ang University\\
{\small\texttt{jeongsoo.choi@kaist.ac.kr, jihk@cau.ac.kr, junseok.ahn@kaist.ac.kr, joonson@kaist.ac.kr}}}

\begin{document}

\maketitle

\begin{abstract}
Cross-lingual dubbing aims to convert source speech into target-language speech that remains synchronized with the original video.
The key to this task is to produce a faithful translation that is naturally speakable within the source timing, which is non-trivial because languages differ in duration, rhythm, and articulation.
In this paper, we propose X-Dubbing, which formulates the task as a joint problem of time-aware speech-to-text translation and video-aware text-to-speech. 
Specifically, our method consists of a Dubbing Translator that generates temporally structured scripts and a Voice Actor that renders them in synchrony with the source video, both grounded in a language-agnostic timing budget. 
We further refine the translator with dubbing-aware feedback from the synthesized speech, encouraging scripts that yield well-synchronized dubbing.
This design requires no paired dubbing data, as the two components are trained on speech-text and monolingual audio-visual data respectively.
Extensive experiments show that X-Dubbing outperforms recent baselines across diverse metrics, including isochrony and audio-visual correspondence.
\end{abstract}

\begin{figure}[!t]
    \centering
    \includegraphics[width=0.98\columnwidth]{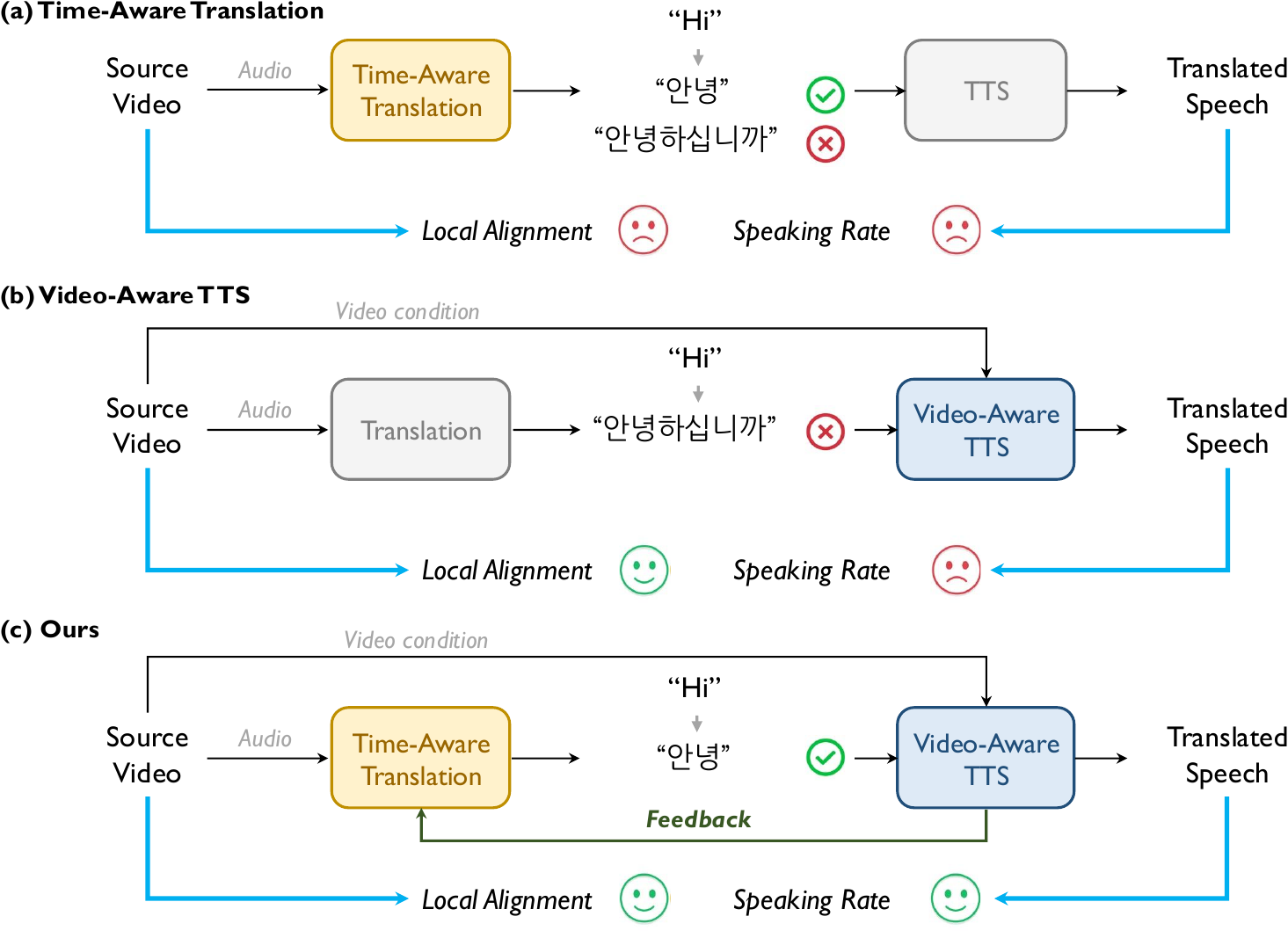}
    \caption{
        \textbf{Comparison of fixed-video dubbing pipelines.}
        (a) Time-aware translation generates timing-constrained scripts for a visually-blind TTS; optimized in isolation, this misses fine-grained visual boundaries, yielding poor local alignment and unnatural speaking rates. 
        (b) Video-aware TTS conditions synthesis on source frames but receives unconstrained scripts from a standard translation module. Local alignment is secured, yet compensating for length mismatches severely distorts the speaking rate. 
        (c) \textbf{\ours{}} (ours) couples time-aware translation with video-aware synthesis via end-to-end feedback. The translation module can now select scripts that inherently satisfy visual constraints, achieving both precise local alignment and a natural speaking rate.
    }
    \label{fig:teaser}
\end{figure}

\section{Introduction}
Cross-lingual dubbing~\cite{federico2020speech, agarwal2023findings} makes video content accessible across languages by replacing the original speech with translated speech.
For the dubbing to feel natural, the translated speech should not only convey the original meaning but also fit the timing and lip movements in the video~\cite{saboo2019integration,chaume2004synchronization}.
In professional dubbing, translators adapt scripts to these constraints, while voice actors deliver them with appropriate timing and articulation~\cite{spiteri2021english, spiteri2023quality}.
These complementary roles highlight the need to consider both the choice of translation and its spoken realization under the constraints of the source video.

One way to address these constraints is to generate translated speech together with corresponding lip movements~\cite{choi2024av2av,cho2025mavflow,chen2026just}.
However, modifying lip movements can introduce inconsistencies with the original scene dynamics, including head movements and body gestures~\cite{yang2025infinitetalk}.
This approach also requires generating a separate video for each target language, while multilingual streaming can reuse the same video with audio tracks in different languages~\cite{sodagar2011mpeg}.
We therefore focus on fixed-video dubbing, where the source video remains unchanged and the translated speech is generated to fit its timing and lip movements.

Analogous to professional dubbing, fixed-video dubbing involves two complementary aspects, \textit{time-aware translation} and \textit{video-aware speech synthesis}.
The former aims to generate semantically valid translations whose expected spoken duration is close to that of the source speech~\cite{lakew2022isometric, chronopoulou2023jointly, wu2023videodubber, cui2025fine}.
The latter converts a given translation into speech while accounting for the visual constraints of the source video~\cite{sahipjohn2024dubwise, wang2025syncvoice}.
However, because text does not explicitly specify speech timing, a translation generated without considering how it will be synthesized may be difficult to render naturally under the video's constraints~\cite{chronopoulou2023jointly, choi2025dub}.
Natural dubbing therefore calls for coordinating translation and speech synthesis based on how the translated content is realized in speech (Fig.~\ref{fig:teaser}).

End-to-end speech-to-speech translation (S2ST) \cite{barrault2023seamless,labiausse2025high} offers a promising way to implicitly couple these two aspects.
These systems, however, do not explicitly model the visual constraints of the source video.
More critically, their supervised adaptation to fixed-video dubbing is limited by the scarcity of paired dubbing data.

In this paper, we propose \ours{}, a fixed-video dubbing framework that tightly integrates \textit{time-aware translation} and \textit{video-aware speech synthesis} through feedback from speech output, without requiring paired cross-lingual dubbing data.
Our framework consists of two main components: a \emph{Dubbing Translator}, which generates translated scripts with explicit pause markers based on source-side timing information, and a \emph{Voice Actor}, which controls speech timing using these markers and frame-level visual cues from the source video.
To coordinate these components, we evaluate candidate translations not only with text-based criteria, but also through the speech synthesized by the Voice Actor.
We use this combined feedback to optimize the Dubbing Translator, encouraging translations that can be rendered naturally under the source video's constraints.

Our contributions are as follows.
(1) We formulate fixed-video cross-lingual dubbing as a problem of coordinating time-aware translation with video-aware speech generation under shared temporal and visual constraints.
(2) We optimize translation by incorporating feedback from the resulting speech, enabling dubbing-aware training without paired dubbing data.
(3) We demonstrate through experiments that our framework achieves strong performance across translation quality, audio-visual synchrony, and perceptual dubbing quality.

\section{Related Work}
\subsection{Time-Aware Machine Translation}
Since dubbing requires translated speech to fit the source timing, early work used isometry, which encourages source-target sentence length matching, as a tractable proxy for isochrony at the translation stage~\cite{federico2020speech,lakew2022isometric}.
Duration-aware methods predict target speech durations to guide translation under source timing constraints~\cite{wu2023videodubber,chronopoulou2023jointly,pal2023improving} or use synthesized speech durations for preference optimization~\cite{cui2025fine}.
Pause-aware methods further consider pause structure in source speech to align the boundaries of corresponding speech segments~\cite{tam2022isochrony,li2024pause}.
However, local alignment within each segment also depends on how the speech generator controls the timing of the translated content.
We therefore couple time-aware translation with a controllable speech generator conditioned on the source video.

\subsection{Video-Aware Speech Generation}
To reduce audio-visual mismatches that can disrupt the viewing experience, several dubbing systems use the source video to guide speech synthesis~\cite{hu2021neural, hassid2022more}.
In monolingual dubbing, where the script is consistent with the articulation shown in the video, synthesis can focus on temporal and prosodic alignment~\cite{cong2023learning,sung2025voicecraft,choi2025aligndit}.
In cross-lingual dubbing, however, translation changes the length and phonetic content of the script, making it harder to fit the original video.
Recent systems extend video-aware speech synthesis to cross-lingual dubbing~\cite{sahipjohn2024dubwise,wang2025syncvoice}, but matching the source timing while keeping the translated script fixed can lead to unnatural speaking rates~\cite{federico2020speech}.
We therefore couple the translator with a video-conditioned speech generator, using feedback from the generated speech to optimize the translator for the source video's temporal and visual constraints.

\begin{figure*}[t]
    \centering
    \includegraphics[width=0.75\textwidth]{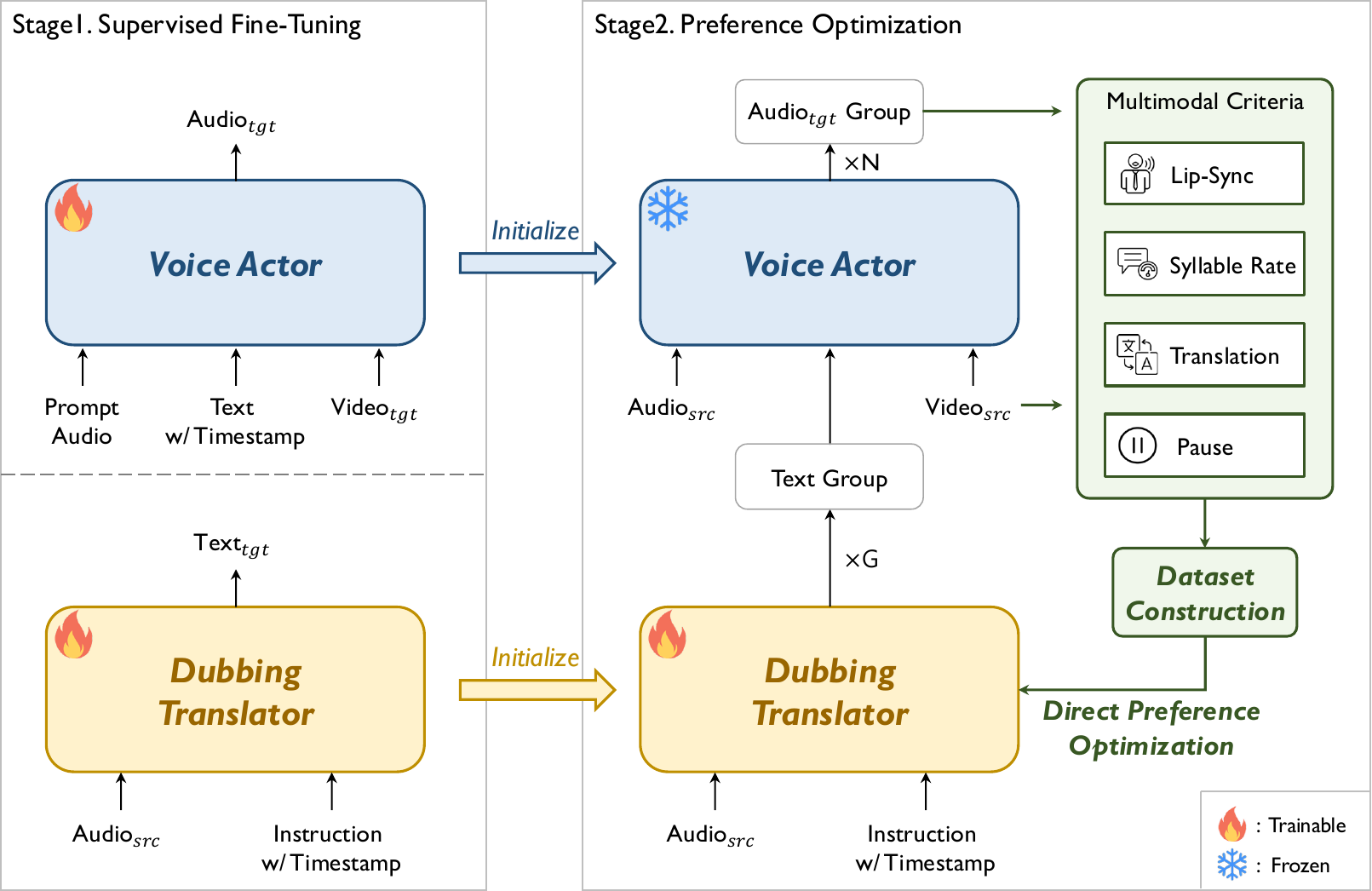}
    \caption{
        \textbf{Method overview of \ours{}.}
        The proposed framework couples a timing-aware dubbing translator and a video-conditioned voice actor through synthesized dubbing feedback.
    }
    \label{fig:method_overview}
\end{figure*}

\subsection{Speech-to-Speech Translation}
Recent end-to-end speech-to-speech translation (S2ST) systems have achieved strong translation quality in generating target-language speech from source speech~\cite{jia2022translatotron,lee2022textless,kim2024textless,seamless2025joint}.
Beyond semantic transfer, expressive S2ST systems can also preserve paralinguistic characteristics of the source speech, such as speaker identity, prosody, and speaking style~\cite{barrault2023seamless, le2024transvip, labiausse2025high}.
These capabilities make S2ST a natural candidate for cross-lingual dubbing, where the target speech should be faithful, natural, and consistent with the source speaker.
Fixed-video dubbing, however, additionally requires the generated speech to follow the source timing and pause structure while aligning with the articulation shown in the video.
Training S2ST systems for these constraints with paired cross-lingual dubbing data is difficult to scale, as such data are scarce.
Our framework instead uses speech–text pairs to train the translator and monolingual audio-visual data to train the speech generator.

\section{Method}
\label{sec:method}
We introduce \ours{}, a framework that converts source speech into target-language dubbed speech that is naturally aligned with the original video. 
As illustrated in Fig.~\ref{fig:method_overview}, the framework consists of a \emph{Dubbing Translator} and a \emph{Voice Actor}.
The Dubbing Translator generates a target-language script with \texttt{<pause>} markers that segment the translated content according to the timing structure of the source speech.
The script is then rendered by the Voice Actor, which uses the source audio as a speaker reference while guiding speech timing with temporal and visual cues.
After supervised fine-tuning, we further optimize the Dubbing Translator using multimodal feedback on candidate translations rendered by the Voice Actor, enabling the model to improve the final dubbing quality.

\subsection{Preliminary}
\paragraph{Speech-to-text translation with multimodal large language models.}
Conventional speech translation often relies on a cascaded pipeline that couples automatic speech recognition (ASR) with machine translation (MT). 
However, passing only the transcript to MT limits access to acoustic cues from the source audio, such as prosody, rhythm, and speaking rate.
Recent multimodal LLMs, such as Qwen2.5-Omni~\cite{xu2025qwen25}, extend decoder-only language models with continuous audio encoders.
Pretrained on web-scale multimodal corpora and aligned via instruction tuning, Qwen2.5-Omni exhibits robust zero-shot performance on speech understanding and translation, and reliably follows natural-language instructions that interleave audio and text within a single prompt. 
Structurally, its Whisper-initialized audio tower~\cite{radford2023robust} encodes source audio directly into acoustic embeddings that the language model reads alongside text tokens.
By bypassing an intermediate textual transcript, this architecture allows the model to generate target-language text directly conditioned on acoustic representations of the source utterance.

\paragraph{Text-to-speech generation with conditional flow-matching.}
Flow matching~\cite{lipman2023flow} trains a generative model to predict a continuous-time vector field that transports a Gaussian prior into the data distribution via ODE integration at inference.
Applied to TTS, the data are mel-spectrogram frames, and the vector field is conditioned on the target text and a short reference utterance from which the speaker identity is transferred~\cite{chen2025f5}. 
The synthesizer we build on, X-Voice~\cite{xu2026x}, instantiates this recipe with a Diffusion Transformer (DiT)~\cite{peebles2023scalable} backbone.
Trained for zero-shot cross-lingual voice cloning, it enables a reference utterance in one language to drive a mel-spectrogram in another while preserving the source speaker's timbre. 
Beyond preserving identity, X-Voice is particularly suited to dubbing through its \emph{length controllability}.
In X-Voice, the output sequence length is determined by the temporal dimension of the initial noise sample, allowing a mel-spectrogram to be generated with the duration required by the source utterance.

\subsection{Dubbing Translator: Time-Aware S2TT}
\label{sec:method:translator}
The Dubbing Translator model is responsible for generating a target-language script that can be naturally spoken under the timing constraints of the source utterance.
This differs from conventional speech-to-text translation, which typically prioritizes fluency and semantic fidelity without explicitly constraining spoken duration.
In dubbing, word choice and segmentation determine how much speech is assigned to each source segment, affecting speaking rate, pause placement, and the quality of the downstream rendered dubbing.
We therefore condition the translator on source-side timing information and train it to produce a target-language script with explicit temporal cues.

Let $A_s$ denote the source-language speech. From $A_s$, we extract a source timing budget using voice activity detection (VAD):
\begin{equation}
\mathcal{B}
=
f_{\mathrm{VAD}}(A_s)
=
\left(D_s, \{[s_k,e_k]\}_{k=1}^{K}\right),
\end{equation}
where $D_s$ is the total audio duration and $\{[s_k,e_k]\}_{k=1}^{K}$ is the ordered sequence of speech segments.
The gaps between adjacent segments define source-side pauses.
This budget provides a coarse, language-agnostic temporal structure of the source utterance, capturing when speech and silence occur.
This makes the same timing representation applicable across languages, sidestepping token-level source-target alignment.

Conditioned on the source speech and its timing budget, the Dubbing Translator $g_\theta$ generates a target-language dubbing script:
\begin{equation}
T = g_\theta(A_s, \mathcal{B}),
\end{equation}
where $T$ consists of target-language words interleaved with \texttt{<pause>} markers.
These markers segment the translated content according to the source pause structure.
The Voice Actor uses this segmentation to guide speech timing during synthesis, as described in Section~\ref{sec:method:voice}.

\paragraph{Isochrony-normalized projection.}
To train $g_\theta$, we need target scripts that contain \texttt{<pause>} markers aligned with the source timing budget.
Ideally, such supervision would come from paired cross-lingual dubbing data, where the target-language script and speech are already adapted to the source utterance. However, such data are difficult to obtain at scale.
We therefore construct timing-aware training targets from ordinary parallel S2TT data, which provide source speech and target text but no target-side timing.

A naive strategy is to project the source pauses directly onto the target sentence according to relative time.
This is problematic because the reference target text is not written under dubbing constraints.
Depending on the translation, the target sentence may contain substantially more or less speakable content than the source utterance.
Using the original source timing budget would therefore impose timing constraints that may not match the expected spoken duration of the projected target script.

We address this mismatch with \emph{isochrony-normalized projection} (INP).
For each training pair, we first estimate how long the target sentence would take to speak using language-normalized syllable counts.
Let $N_s$ and $N_t$ denote the syllable counts of the source transcript and the target translation, respectively.
Given language-level average syllable rates $\bar{r}_s$ and $\bar{r}_t$~\cite{pellegrino2011cross}, we compute the expected target duration as
\begin{equation}
\hat{D}_t
=
D_s
\cdot
\frac{N_t / \bar{r}_t}{N_s / \bar{r}_s},
\end{equation}
where $D_s$ is the source audio duration.
We then stretch or shrink the source timing budget from $D_s$ to $\hat{D}_t$, preserving the relative positions of the source speech segments and pauses.
To construct $T^{\mathrm{sft}}$, we allocate target syllables in proportion to the source speech-segment durations and insert \texttt{<pause>} markers at the nearest word boundaries.

\paragraph{Supervised fine-tuning.}
We instantiate the translator model $g_\theta$ with Qwen2.5-Omni~\cite{xu2025qwen25}.
The model receives the source speech together with the rescaled timing budget, including the total duration, VAD-derived speech intervals, and intervening pauses.
It is then fine-tuned to generate the constructed script $T^{\mathrm{sft}}$ interleaved with \texttt{<pause>} markers.
This rescaling applies only to the SFT timing instruction, while preference optimization and inference use the original source timing budget $\mathcal{B}$.

This supervised stage initializes the translator with a stable timing-aware generation interface before preference optimization.
We train LoRA adapters~\cite{hu2022lora} on both the audio encoder and the language model, and also fine-tune the audio projection module.
The resulting translator is later refined using feedback from synthesized dubbing in Section~\ref{sec:method:dpo}.

\subsection{Voice Actor: Video-Aware TTS}
\label{sec:method:voice}
A pause-marked script defines the textual structure of the target utterance, but dubbing requires this structure to be grounded on the acoustic timeline of the source clip.
Even with a fixed total duration, the generated speech can still be locally misaligned with the source timeline.
We therefore introduce source-derived frame-level conditioning to guide the realization of the script.

Given a target script $T$, source speech $A_s$ as a speaker reference, source timing budget $\mathcal{B}$, and source video $V_s$, the Voice Actor $h_\phi$ generates target-language speech:
\begin{equation}
A_t = h_\phi(T, A_s, \mathcal{B}, V_s).
\end{equation}
We build $h_\phi$ on X-Voice~\cite{xu2026x} and adapt it to use two complementary forms of source conditioning.
The VAD-derived cue provides a frame-level speech activity plan, while the SyncNet~\cite{chung2016syncnet}-derived cue provides frame-level visual articulation information.
Together, they allow the Voice Actor to synthesize the translated script under both temporal and visual guidance.

\paragraph{Pause conditioning.}
VAD-derived speech activity provides a lightweight yet informative representation of dubbing isochrony~\cite{le2024transvip}, indicating speech and silence over time. 
We encode it as a binary frame sequence $\mathbf{v}\in\{0,1\}^{L}$, where $L$ is the number of mel-spectrogram frames, $1$ denotes speech, and $0$ denotes silence.
During training, $\mathbf{v}$ is extracted from target speech, whereas at inference it is derived from source speech to transfer the source clip's pause structure to the generated dubbing.

At inference, we set the output mel-spectrogram length to match the source duration and apply $\mathbf{v}$ at the corresponding acoustic frames.
On the text side, we derive speech and pause states from the script's \texttt{<pause>} markers and add their embeddings to the text representations.
Together, these cues guide how script segments are realized within the source speech intervals.

\paragraph{Video conditioning.}
Dubbing also requires the generated speech to agree with the visible mouth motion of the source video.
Although lip-reading features and self-supervised phonetic visual representations have been used in monolingual dubbing, they can introduce a training-inference mismatch in cross-lingual settings.
We therefore use SyncNet embeddings as a visual synchrony cue, which provide frame-level information about audio-visual alignment without explicitly imposing source-language linguistic content.
We resample the visual features to the mel-spectrogram frame rate so that the visual and pause cues align with the same output frames.

We inject the aligned temporal and visual cues at both the acoustic input and each DiT block.
At the input, we separately project the cues and add their contributions to the acoustic conditioning.
At each block, the two cues independently predict frame-wise scale and bias terms, which are summed for FiLM modulation~\cite{perez2018film},
\begin{equation}
\mathrm{FiLM}(\mathbf{h}^{\ell}_i)
=
(1+\gamma^{\ell}_{p,i}+\gamma^{\ell}_{v,i})
\odot\mathbf{h}^{\ell}_i
+\beta^{\ell}_{p,i}+\beta^{\ell}_{v,i},
\end{equation}
where $\mathbf{h}^{\ell}_i$ denotes the hidden representation at frame $i$ entering DiT block $\ell$, and the subscripts $p$ and $v$ denote the pause and visual conditions, respectively.
This dual injection provides both early access and persistent frame-level control.

\paragraph{Condition dropout.}
During training, the target speech and conditioning video are in the same language, whereas inference is cross-lingual.
We therefore independently drop the pause and visual conditions with probability $0.5$ to reduce over-reliance on either cue and improve robustness to this training-inference mismatch.

\subsection{Preference Optimization with End-to-End Feedback}
\label{sec:method:dpo}
During supervised fine-tuning, the Dubbing Translator learns to produce timing-aware scripts but receives no feedback on how its word choices and segmentation affect the synthesized speech.
We therefore render candidate translations with the frozen Voice Actor and evaluate both the scripts and their spoken realizations.
We use these evaluations to construct preference pairs for further optimizing the Dubbing Translator.

\paragraph{Preference data construction.}
Starting from the SFT translator, we generate $G$ candidate translations for each source utterance using its original timing budget.
Each candidate is rendered by the fine-tuned Voice Actor with $N$ random seeds and evaluated using four multimodal criteria.
First, we compare the script's implied segment-level syllable rates with those of the source, using source segment durations and language-level rate normalization.
Second, we measure reference-free translation quality with a quality estimation model~\cite{rei2022cometkiwi}.
Third, we measure lip synchronization between the synthesized speech and source video using the SyncNet-based LSE-D and LSE-C metrics~\cite{prajwal2020lip}.
We average the lip-sync scores across the $N$ renditions to reduce sensitivity to synthesis randomness.
Finally, we check pause consistency by requiring the number of \texttt{<pause>} markers to match the number of pauses in the source timing budget, discarding candidates that violate this constraint.
Among candidates satisfying the pause constraint, we construct preference pairs using Pareto dominance, preferring a candidate only when it performs no worse on any scored criterion and better on at least one.

\paragraph{Direct preference optimization.}
We fine-tune the Dubbing Translator on these fixed preference pairs using the direct preference optimization (DPO) objective~\cite{rafailov2023direct}:
\begin{equation}
\mathcal{L}_{\mathrm{DPO}}
=
-\log \sigma(\beta \Delta),
\end{equation}
where
\begin{equation}
\Delta
=
\log \frac{\pi_\theta(y_c \mid x)}{\pi_{\mathrm{ref}}(y_c \mid x)}
-
\log \frac{\pi_\theta(y_r \mid x)}{\pi_{\mathrm{ref}}(y_r \mid x)}.
\end{equation}
Here, $x=(A_s,\mathcal{B})$ denotes the source speech and timing instruction, while $y_c$ and $y_r$ are the chosen and rejected scripts, respectively.
$\pi_\theta$ denotes the conditional script distribution of $g_\theta$, and $\pi_{\mathrm{ref}}$ is the frozen SFT reference model.
The coefficient $\beta$ controls regularization toward this reference.
As in supervised fine-tuning, we update the LoRA adapters and audio projection module of the Dubbing Translator while keeping the remaining parameters fixed.

\begin{table*}[t]
\centering
\small
\setlength{\tabcolsep}{3pt}
\resizebox{0.95\textwidth}{!}{
\begin{tabular}{l cc c cccc cc c}
\toprule
& \multicolumn{2}{c}{Translation Quality}
& {Speaker Similarity}
& \multicolumn{4}{c}{Temporal Alignment}
& \multicolumn{2}{c}{Lip Sync}
& {S. Qual.} \\
\cmidrule(lr){2-3} \cmidrule(lr){4-4} \cmidrule(lr){5-8} \cmidrule(lr){9-10} \cmidrule(lr){11-11}
Method
& ASR-BLEU$\uparrow$ & ASR-chrF++$\uparrow$
& SpkSim$\uparrow$
& SyllRate $\rho\uparrow$ & Pause IoU$\uparrow$ & SLC@0.2$\uparrow$ & SLC@0.4$\uparrow$
& LSE-C$\uparrow$ & LSE-D$\downarrow$
& UTMOS$\uparrow$ \\
\midrule
AV2AV-Speech
& 33.53 & 56.29
& 0.212
& 0.082 & 0.111 & 72.2 & 94.7
& 1.56 & 13.10
& 3.121 \\
SeamlessExpressive
& 38.44 & 60.27
& 0.280
& 0.518 & 0.166 & 70.6 & 97.1
& 1.62 & 12.31
& 3.433 \\
Hibiki
& 37.35 & 60.03
& 0.358
& 0.295 & 0.307 & 75.3 & 94.1
& 1.48 & 12.62
& \textbf{3.864} \\
Dub-S2ST
& 23.69 & 47.51
& 0.282
& 0.211 & 0.106 & 68.2 & 92.4
& 1.56 & 12.91
& 3.628 \\
\ours
& \textbf{42.34} & \textbf{64.06}
& \textbf{0.497}
& \textbf{0.617} & \textbf{0.551} & \textbf{97.1} & \textbf{100.0}
& \textbf{2.58} & \textbf{11.30}
& 3.748 \\
\bottomrule
\end{tabular}
}
\vspace{-2mm}
\caption{Main results on the mTEDx Fr$\rightarrow$En test set.}
\label{tab:main_results}
\end{table*}

\section{Experiment}
\label{sec:exp}
\subsection{Datasets}
\label{sec:exp:data}
We use French speech paired with English translations from the mTEDx training split~\cite{salesky2021multilingual} for supervised fine-tuning of the Dubbing Translator.
After preprocessing, this subset contains 11,092 clips totaling 14.9 hours, with durations ranging from 2 to 10 seconds.
For supervised fine-tuning of the Voice Actor, we use 95,474 English audio-visual clips with corresponding transcripts from LRS3~\cite{afouras2018lrs3}, totaling approximately 200 hours.
For subsequent preference optimization of the Dubbing Translator, we use 7,752 preference pairs constructed from the \mbox{mTEDx} training subset.
We evaluate our framework on a preprocessed subset of the mTEDx French-to-English test split.

\subsection{Evaluation Metrics}
\label{sec:exp:metrics}
We evaluate the generated speech along the dimensions of translation quality, speaker similarity, temporal alignment, lip sync, and speech quality using objective metrics. We complement these with subjective metrics obtained through a human study.

\subsubsection{Objective Metrics}
\paragraph{Translation quality.}
We report \textbf{ASR-BLEU} and \textbf{ASR-chrF++} by comparing Whisper-medium~\cite{radford2023robust}
transcriptions of the generated speech with the reference translations, using
SacreBLEU~\cite{post2018call} and chrF++~\cite{popovic2017chrf++}, respectively.

\paragraph{Speaker similarity.}
We measure \textbf{SpkSim} as the cosine similarity between speaker embeddings of the source and generated speech, extracted using a WavLM-based ECAPA-TDNN~\cite{chen2022wavlm}.

\paragraph{Temporal alignment.}
We use Silero VAD \cite{silerovad} to obtain active speech durations and pause intervals for the source and generated speech.
\textbf{SyllRate $\rho$}~\cite{barrault2023seamless} measures the Spearman correlation between source and generated syllable rates (syllables per second of active speech).
\textbf{Pause IoU} averages the intersection over union of pause intervals over clips containing source pauses.
Speech Length Compliance (\textbf{SLC@$p$})~\cite{wu2023videodubber} measures the percentage of clips whose active speech duration is between $(1-p)$ and $(1+p)$ times that of the source, where $p\in\{0.2,0.4\}$.

\paragraph{Lip synchronization.}
We compute Lip Sync Error Confidence (\textbf{LSE-C}) and Lip Sync Error Distance (\textbf{LSE-D})~\cite{prajwal2020lip} between the generated speech and the source video using pretrained SyncNet~\cite{chung2016syncnet}.

\paragraph{Speech quality.}
We assess the perceptual quality of the generated speech using \textbf{UTMOS}~\cite{saeki2022utmos}.

\subsubsection{Subjective Metrics}
To complement the objective metrics, we conduct a mean opinion score (MOS) evaluation with 15 raters on 12 test clips.
For each clip, the same source audio is dubbed into the target language by our framework and each baseline, and raters score each resulting dubbed speech against the original source video on a five-point scale from 1 (poor) to 5 (excellent): \textit{Audio-Visual Correspondence}, evaluating how well the speech matches the lip movements and timing in the video, and \textit{Overall Naturalness}, evaluating how naturally the speech is delivered in terms of speaking rate, pauses, and rhythm.

\subsection{Implementation Details}
The Dubbing Translator and Voice Actor are initialized from Qwen2.5-Omni-3B~\cite{xu2025qwen25} and X-Voice~\cite{xu2026x}, respectively.
We apply LoRA~\cite{hu2022lora} with rank $8$ and $\alpha = 32$ to linear layers in both models.
The audio projection module of the Dubbing Translator and the added pause and visual conditioning modules of the Voice Actor are fully trainable.
We use Silero VAD~\cite{silerovad} for speech activity detection, with a speech probability threshold of $0.5$ and a minimum silence duration of $100$\,ms.
Word-level alignments for LRS3 are obtained using the UnitY2 aligner~\cite{barrault2023seamless} in stopes~\cite{andrews2022stopes} library.
We synthesize French reference speech with pretrained X-Voice, using the corresponding LRS3 audio as the speaker reference.

\subsection{Baseline Models}
\label{sec:exp:baselines}
We compare \ours{} against recent open-source systems, each addressing a distinct aspect of cross-lingual dubbing.
AV2AV~\cite{choi2024av2av} translates audio-visual speech into synchronized target speech and video. For our fixed-video setting, we evaluate its speech output against the original source video and refer to this baseline as {AV2AV-Speech}, following prior work~\cite{goncalves2025improving}.
SeamlessExpressive~\cite{barrault2023seamless} translates speech while preserving source prosody and vocal style.
Hibiki~\cite{labiausse2025high} learns speech and pause timing from contextually aligned training data for simultaneous S2ST.
Dub-S2ST~\cite{choi2025dub} uses explicit duration control and speech unit-based speaking rate adaptation for textless S2ST.

\begin{table}[t]
\centering
\small
\resizebox{\columnwidth}{!}{
\begin{tabular}{l c c cc c}
\toprule
Method
& ASR-BLEU$\uparrow$
& Pause IoU$\uparrow$
& LSE-C$\uparrow$ & LSE-D$\downarrow$
& UTMOS$\uparrow$ \\
\midrule
Cascaded        & 41.93 & 0.129 & 1.61 & 12.94 & 3.617 \\
Segment-level   & 38.61 & \textbf{0.578} & 1.51 & 12.45 & 3.552 \\
Re-ranking      & 41.35 & 0.149 & 2.11 & 12.41 & 3.600 \\
Paraphrase      & 26.81 & 0.139 & 2.30 & 12.19 & 3.596 \\
\ours           & \textbf{42.34} & 0.551 & \textbf{2.58} & \textbf{11.30} & \textbf{3.748} \\
\bottomrule
\end{tabular}
}
\vspace{-2mm}
\caption{Comparison with cascaded baselines.}
\label{tab:comparison_cascade}
\end{table}

\section{Experimental Results}
\label{sec:results}
\subsection{Main Results}
\label{sec:results:main}
Table~\ref{tab:main_results} compares \ours{} with baseline systems in the fixed-video dubbing setting.
While prior systems show strengths in translation quality, speaker similarity, or speech quality, they remain limited in satisfying dubbing-specific constraints such as pause structure, speaking-rate consistency, and lip synchrony.
In contrast, \ours{} achieves the best temporal alignment and lip-sync scores, while also attaining the highest translation and speaker similarity scores and maintaining competitive speech quality.

\subsection{Comparison with Cascaded Baselines}
\label{sec:results:cascade}
Table~\ref{tab:comparison_cascade} compares \ours{} with alternative cascading strategies built on the same pretrained translation and speech synthesis models.
The vanilla cascade provides strong translation quality but limited pause alignment, whereas segment-level translation improves pause alignment with lower translation quality.
Inspired by prior work on candidate search and paraphrasing for dubbing~\cite{saragih2025beyond,hong2026ps}, we implement re-ranking and paraphrase-based search by generating $G = 8$ text candidates and synthesizing speech from each candidate with $N = 4$ random seeds.
We apply min-max scaling to the lip-sync (LSE-C $-$ LSE-D) and reference-free translation quality (COMET-Kiwi~\cite{rei2022cometkiwi}) scores, then select the speech output with the highest weighted sum using weights of 1.6 and 0.4, respectively.
These searches improve lip synchrony but require repeated synthesis and scoring, whereas \ours{} achieves the best performance in translation quality and lip synchrony without candidate search at inference.

\begin{table}[t]
\centering
\small
\resizebox{\columnwidth}{!}{
\begin{tabular}{l c c cc}
\toprule
Method
& ASR-BLEU$\uparrow$
& SyllRate $\rho\uparrow$
& Pause IoU$\uparrow$
& SLC@0.2$\uparrow$ \\
\midrule
\ours           & \textbf{42.34} & \textbf{0.617} & 0.551 & \textbf{97.1} \\
\ours~(SFT)     & 41.55 & 0.577 & \textbf{0.563} & 94.1 \\
\quad w/o INP   & 41.49 & 0.538 & 0.554 & 92.9 \\
\bottomrule
\end{tabular}
}
\vspace{-2mm}
\caption{Ablation study of training components for the Dubbing Translator.}
\label{tab:ablation_s2tt}
\end{table}

\begin{table}[t]
\centering
\small
\setlength{\tabcolsep}{3pt}
\resizebox{\columnwidth}{!}{
\begin{tabular}{l ccccc}
\toprule
Method
& Seg. WER$\downarrow$
& Pause IoU$\uparrow$
& LSE-C$\uparrow$
& LSE-D$\downarrow$
& UTMOS$\uparrow$ \\
\midrule
\ours~(SFT)                     & \textbf{13.97} & 0.563 & \textbf{2.56} & \textbf{11.30} & 3.741 \\
\quad w/o Synthetic references  & 16.60 & 0.520 & 2.38 & 11.46 & 3.771 \\
\quad w/o Frame-level FiLM      & 16.73 & 0.386 & 1.91 & 12.04 & 3.755 \\
\quad w/o Text-side pause emb.  & 15.37 & \textbf{0.573} & 2.52 & \textbf{11.30} & 3.708 \\
\quad w/o Pause conditioning    & 19.23 & 0.366 & 2.47 & 11.33 & 3.724 \\
\quad w/o Video conditioning    & 14.45 & 0.436 & 1.55 & 12.62 & \textbf{3.819} \\
\bottomrule
\end{tabular}
}
\vspace{-2mm}
\caption{Ablation study of the components of the Voice Actor.}
\label{tab:ablation_tts}
\end{table}

\subsection{Ablation Study}
\subsubsection{Dubbing Translator}
\label{sec:results:ablation_s2tt}
Table~\ref{tab:ablation_s2tt} examines how INP and preference optimization contribute to the Dubbing Translator.
The DPO-trained translator outperforms the SFT model in translation quality, speaking-rate consistency, and duration compliance.
Removing INP from the SFT model weakens pacing control, supporting the use of normalized timing supervision during supervised fine-tuning.
Together, these results suggest that INP-based supervision and dubbing-aware preference optimization help the translator produce scripts that better fit the source timing.

\subsubsection{Voice Actor}
\label{sec:results:ablation_tts}
Table~\ref{tab:ablation_tts} examines the components that make the Voice Actor controllable.
We report segment-level word error rate (Seg. WER) by comparing ASR transcriptions of generated speech within source speech intervals with the corresponding script segments.
Removing frame-level FiLM substantially weakens both pause alignment and lip synchrony, supporting the use of temporal and visual conditioning within the DiT blocks.
Removing pause conditioning increases Seg. WER and weakens pause alignment, while removing video conditioning degrades both pause alignment and lip synchrony.
These results support combining pause and video conditioning through frame-level FiLM to align synthesized speech with the source timing and lip movements.

\begin{table}[t]
\centering
\small
\resizebox{\columnwidth}{!}{
\begin{tabular}{l cc}
\toprule
Method
& Audio-Visual Corr.
& Overall Naturalness \\
\midrule
AV2AV-Speech        & 2.32 $\pm$ 0.16 & 2.75 $\pm$ 0.16 \\
SeamlessExpressive  & 2.58 $\pm$ 0.18 & 3.05 $\pm$ 0.15 \\
Hibiki              & 2.64 $\pm$ 0.21 & 2.52 $\pm$ 0.16 \\
Dub-S2ST            & 2.05 $\pm$ 0.16 & 2.40 $\pm$ 0.15 \\
Cascaded            & 3.13 $\pm$ 0.17 & 3.54 $\pm$ 0.14 \\
\ours               & \textbf{4.01 $\pm$ 0.12} & \textbf{3.99 $\pm$ 0.13} \\
\bottomrule
\end{tabular}
}
\vspace{-2mm}
\caption{MOS evaluation.}
\label{tab:mos}
\end{table}

\subsection{Human Evaluation}
Table~\ref{tab:mos} reports human evaluation results. \ours{} achieves the highest ratings in both audio-visual correspondence and overall naturalness, confirming that the improvements in automatic timing and lip-sync metrics translate to perceived dubbing quality.

\section{Conclusion}
We presented X-Dubbing, a cross-lingual dubbing framework that treats translation and speech synthesis as a coordinated process rather than isolated stages. 
By connecting a time-aware translator and a video-aware speech generator through a language-agnostic timing budget and refining the translator with feedback from synthesized speech, X-Dubbing generates natural dubbed speech synchronized with the source video, without requiring paired dubbing data.
Extensive evaluations confirm improvements in translation quality, isochrony, lip-sync, and perceived dubbing quality. More broadly, our results suggest that downstream speech feedback offers a scalable path toward dubbing-aware translation.

\section*{Limitations}
\label{sec:limitations}
Our experiments focus on French-to-English monologue videos with a single visible speaker.
Extending X-Dubbing to more diverse language pairs, multi-speaker conversations, and complex visual scenes remains an important direction for future work.
In addition, as with other generative approaches, the quality of the dubbed speech may depend on the characteristics of the source audio and video.

\section*{Ethical Considerations}
As X-Dubbing can reproduce a speaker’s voice in another language, it may be misused to generate misleading content.
We encourage responsible use with the speaker’s consent and clear disclosure of synthetic speech.
All datasets used in this work are publicly available and used in accordance with their licenses, and human evaluation was conducted with the informed consent of participants.

\bibliography{custom}

@string{aaai     = "Proc. AAAI"}

@string{accvw    = "ACCV Workshops"}

@string{acl      = "Proc. ACL"}

@string{acmmm    = "Proc. ACM MM"}

@string{apsipa   = "Proc. APSIPA ASC"}

@string{cvpr     = "Proc. CVPR"}

@string{emnlpd   = "Proc. EMNLP: System Demonstrations"}

@string{emnlpf   = "Findings of EMNLP"}

@string{icassp   = "Proc. ICASSP"}

@string{iccv     = "Proc. ICCV"}

@string{iclr     = "Proc. ICLR"}

@string{icml     = "Proc. ICML"}

@string{icpr     = "Proc. ICPR"}

@string{is       = "Proc. Interspeech"}

@string{iwslt    = "Proc. IWSLT"}

@string{naacl    = "Proc. NAACL"}

@string{neurips  = "NeurIPS"}

@string{siggraph = "Proc. ACM SIGGRAPH"}

@string{wmt      = "Proc. WMT"}

@string{jstsp    = "IEEE Journal of Selected Topics in Signal Processing"}

@string{taslp    = "IEEE/ACM Trans. on Audio, Speech, and Language Processing"}

@article{afouras2018lrs3,
  title={{LRS3-TED}: a large-scale dataset for visual speech recognition},
  author={Afouras, Triantafyllos and Chung, Joon Son and Zisserman, Andrew},
  journal={arXiv preprint arXiv:1809.00496},
  year={2018}
}

@inproceedings{agarwal2023findings,
  title={Findings of the {IWSLT} 2023 evaluation campaign},
  author={Agarwal, Milind and Agrawal, Sweta and Anastasopoulos, Antonios and Bentivogli, Luisa and Bojar, Ond{\v{r}}ej and Borg, Claudia and Carpuat, Marine and Cattoni, Roldano and Cettolo, Mauro and Chen, Mingda and others},
  booktitle=iwslt,
  year={2023}
}

@inproceedings{andrews2022stopes,
  title={{stopes} - Modular Machine Translation Pipelines},
  author={Andrews, Pierre and Wenzek, Guillaume and Heffernan, Kevin and {\c{C}}elebi, Onur and Sun, Anna and Kamran, Ammar and Guo, Yingzhe and Mourachko, Alexandre and Schwenk, Holger and Fan, Angela},
  booktitle=emnlpd,
  year={2022}
}

@article{barrault2023seamless,
  title={{Seamless}: Multilingual Expressive and Streaming Speech Translation},
  author={Barrault, Lo{\"\i}c and Chung, Yu-An and Meglioli, Mariano Coria and Dale, David and Dong, Ning and Duppenthaler, Mark and Duquenne, Paul-Ambroise and Ellis, Brian and Elsahar, Hady and Haaheim, Justin and others},
  journal={arXiv preprint arXiv:2312.05187},
  year={2023}
}

@incollection{chaume2004synchronization,
  title={Synchronization in dubbing: A translational approach},
  author={Chaume, Frederic},
  booktitle={Topics in Audiovisual Translation},
  year={2004},
  publisher={John Benjamins Publishing Company}
}

@inproceedings{chen2026just,
  title={{Just-Dub-It}: Video dubbing via joint audio-visual diffusion},
  author={Chen, Anthony and Ken Korem, Naomi and Halperin, Tavi and Ben Yosef, Matan and Jelercic, Urska and Bibi, Ofir and Patashnik, Or and Cohen-Or, Daniel},
  booktitle=siggraph,
  year={2026}
}

@article{chen2022wavlm,
  title={{WavLM}: Large-scale self-supervised pre-training for full stack speech processing},
  author={Chen, Sanyuan and Wang, Chengyi and Chen, Zhengyang and Wu, Yu and Liu, Shujie and Chen, Zhuo and Li, Jinyu and Kanda, Naoyuki and Yoshioka, Takuya and Xiao, Xiong and others},
  journal=jstsp,
  year={2022}
}

@inproceedings{chen2025f5,
  title={{F5-TTS}: A fairytaler that fakes fluent and faithful speech with flow matching},
  author={Chen, Yushen and Niu, Zhikang and Ma, Ziyang and Deng, Keqi and Wang, Chunhui and Zhao, Jian and Yu, Kai and Chen, Xie},
  booktitle=acl,
  year={2025}
}

@inproceedings{cho2025mavflow,
  title={{MAVFlow}: Preserving Paralinguistic Elements with Conditional Flow Matching for Zero-Shot {AV2AV} Multilingual Translation},
  author={Cho, Sungwoo and Choi, Jeongsoo and Kim, Sungnyun and Yun, Se-Young},
  booktitle=iccv,
  year={2025}
}

@inproceedings{choi2025dub,
  title={{Dub-S2ST}: Textless Speech-to-Speech Translation for Seamless Dubbing},
  author={Choi, Jeongsoo and Kim, Jaehun and Chung, Joon Son},
  booktitle=emnlpf,
  year={2025}
}

@inproceedings{choi2025aligndit,
  title={{AlignDiT}: Multimodal Aligned Diffusion Transformer for Synchronized Speech Generation},
  author={Choi, Jeongsoo and Kim, Ji-Hoon and Sung-Bin, Kim and Oh, Tae-Hyun and Chung, Joon Son},
  booktitle=acmmm,
  year={2025}
}

@inproceedings{choi2024av2av,
  title={{AV2AV}: Direct Audio-Visual Speech to Audio-Visual Speech Translation with Unified Audio-Visual Speech Representation},
  author={Choi, Jeongsoo and Park, Se Jin and Kim, Minsu and Ro, Yong Man},
  booktitle=cvpr,
  year={2024}
}

@article{chronopoulou2023jointly,
  title={Jointly optimizing translations and speech timing to improve isochrony in automatic dubbing},
  author={Chronopoulou, Alexandra and Thompson, Brian and Mathur, Prashant and Virkar, Yogesh and Lakew, Surafel M and Federico, Marcello},
  journal={arXiv preprint arXiv:2302.12979},
  year={2023}
}

@inproceedings{chung2016syncnet,
  title={Out of time: automated lip sync in the wild},
  author={Chung, Joon Son and Zisserman, Andrew},
  booktitle=accvw,
  year={2016}
}

@inproceedings{cong2023learning,
  title={Learning to dub movies via hierarchical prosody models},
  author={Cong, Gaoxiang and Li, Liang and Qi, Yuankai and Zha, Zheng-Jun and Wu, Qi and Wang, Wenyu and Jiang, Bin and Yang, Ming-Hsuan and Huang, Qingming},
  booktitle=cvpr,
  year={2023}
}

@inproceedings{cui2025fine,
  title={Fine-grained video dubbing duration alignment with segment supervised preference optimization},
  author={Cui, Chaoqun and Huang, Liangbin and Wang, Shijing and Tong, Zhe and Huang, Zhaolong and Zeng, Xiao and Liu, Xiaofeng},
  booktitle=acl,
  year={2025}
}

@inproceedings{federico2020speech,
  title={From speech-to-speech translation to automatic dubbing},
  author={Federico, Marcello and Enyedi, Robert and Barra-Chicote, Roberto and Giri, Ritwik and Isik, Umut and Krishnaswamy, Arvindh and Sawaf, Hassan},
  booktitle=iwslt,
  year={2020}
}

@inproceedings{goncalves2025improving,
  title={Improving lip-synchrony in direct audio-visual speech-to-speech translation},
  author={Goncalves, Lucas and Mathur, Prashant and Niu, Xing and Lavania, Chandrashekhar and Houston, Brady and Vishnubhotla, Srikanth and Sun, Lijia and Ferritto, Anthony},
  booktitle=icassp,
  year={2025}
}

@inproceedings{hassid2022more,
  title={More than words: In-the-wild visually-driven prosody for text-to-speech},
  author={Hassid, Michael and Ramanovich, Michelle Tadmor and Shillingford, Brendan and Wang, Miaosen and Jia, Ye and Remez, Tal},
  booktitle=cvpr,
  year={2022}
}

@inproceedings{hong2026ps,
  title={{PS-TTS}: Phonetic Synchronization in Text-to-Speech for Achieving Natural Automated Dubbing},
  author={Hong, Changi and Song, Yoonah and Park, Hwayoung and Bang, Chaewoon and Ku, Dayeon and Lee, Do Hyun and Kim, Hong Kook},
  booktitle=icpr,
  year={2026}
}

@inproceedings{hu2021neural,
  title={{Neural Dubber}: Dubbing for videos according to scripts},
  author={Hu, Chenxu and Tian, Qiao and Li, Tingle and Wang, Yuping and Wang, Yuxuan and Zhao, Hang},
  booktitle=neurips,
  year={2021}
}

@inproceedings{hu2022lora,
  title={{LoRA}: Low-Rank Adaptation of Large Language Models},
  author={Hu, Edward J and Shen, Yelong and Wallis, Phillip and Allen-Zhu, Zeyuan and Li, Yuanzhi and Wang, Shean and Wang, Lu and Chen, Weizhu},
  booktitle=iclr,
  year={2022}
}

@inproceedings{jia2022translatotron,
  title={Translatotron 2: High-quality direct speech-to-speech translation with voice preservation},
  author={Jia, Ye and Ramanovich, Michelle Tadmor and Remez, Tal and Pomerantz, Roi},
  booktitle=icml,
  year={2022}
}

@article{kim2024textless,
  title={Textless Unit-to-Unit Training for Many-to-Many Multilingual Speech-to-Speech Translation},
  author={Kim, Minsu and Choi, Jeongsoo and Kim, Dahun and Ro, Yong Man},
  journal=taslp,
  year={2024}
}

@inproceedings{labiausse2025high,
  title={High-fidelity simultaneous speech-to-speech translation},
  author={Labiausse, Tom and Mazar{\'e}, Laurent and Grave, Edouard and P{\'e}rez, Patrick and D{\'e}fossez, Alexandre and Zeghidour, Neil},
  booktitle=icml,
  year={2025}
}

@inproceedings{lakew2022isometric,
  title={Isometric {MT}: Neural machine translation for automatic dubbing},
  author={Lakew, Surafel M and Virkar, Yogesh and Mathur, Prashant and Federico, Marcello},
  booktitle=icassp,
  year={2022}
}

@inproceedings{le2024transvip,
  title={{TransVIP}: Speech to speech translation system with voice and isochrony preservation},
  author={Le, Chenyang and Qian, Yao and Wang, Dongmei and Zhou, Long and Liu, Shujie and Wang, Xiaofei and Yousefi, Midia and Qian, Yanmin and Li, Jinyu and Zhao, Sheng and others},
  booktitle=neurips,
  year={2024}
}

@inproceedings{lee2022textless,
  title={Textless speech-to-speech translation on real data},
  author={Lee, Ann and Gong, Hongyu and Duquenne, Paul-Ambroise and Schwenk, Holger and Chen, Peng-Jen and Wang, Changhan and Popuri, Sravya and Adi, Yossi and Pino, Juan and Gu, Jiatao and others},
  booktitle=naacl,
  year={2022}
}

@inproceedings{li2024pause,
  title={Pause-aware automatic dubbing using {LLM} and voice cloning},
  author={Li, Yuang and Guo, Jiaxin and Zhang, Min and Ma, Miaomiao and Rao, Zhiqiang and Zhang, Weidong and He, Xianghui and Wei, Daimeng and Yang, Hao},
  booktitle=iwslt,
  year={2024}
}

@inproceedings{lipman2023flow,
  title={Flow Matching for Generative Modeling},
  author={Lipman, Yaron and Chen, Ricky TQ and Ben-Hamu, Heli and Nickel, Maximilian and Le, Matthew},
  booktitle=iclr,
  year={2023}
}

@inproceedings{pal2023improving,
  title={Improving isochronous machine translation with target factors and auxiliary counters},
  author={Pal, Proyag and Thompson, Brian and Virkar, Yogesh and Mathur, Prashant and Chronopoulou, Alexandra and Federico, Marcello},
  booktitle=is,
  year={2023}
}

@inproceedings{peebles2023scalable,
  title={Scalable diffusion models with transformers},
  author={Peebles, William and Xie, Saining},
  booktitle=iccv,
  year={2023}
}

@article{pellegrino2011cross,
  title={A cross-language perspective on speech information rate},
  author={Pellegrino, Fran{\c{c}}ois and Coup{\'e}, Christophe and Marsico, Egidio},
  journal={Language},
  year={2011}
}

@inproceedings{perez2018film,
  title={{FiLM}: Visual reasoning with a general conditioning layer},
  author={Perez, Ethan and Strub, Florian and De Vries, Harm and Dumoulin, Vincent and Courville, Aaron},
  booktitle=aaai,
  year={2018}
}

@inproceedings{popovic2017chrf++,
  title={{chrF++}: words helping character n-grams},
  author={Popovi{\'c}, Maja},
  booktitle=wmt,
  year={2017}
}

@inproceedings{post2018call,
  title={A Call for Clarity in Reporting {BLEU} Scores},
  author={Post, Matt},
  booktitle=wmt,
  year={2018}
}

@inproceedings{prajwal2020lip,
  title={A lip sync expert is all you need for speech to lip generation in the wild},
  author={Prajwal, KR and Mukhopadhyay, Rudrabha and Namboodiri, Vinay P and Jawahar, CV},
  booktitle=acmmm,
  year={2020}
}

@inproceedings{radford2023robust,
  title={Robust speech recognition via large-scale weak supervision},
  author={Radford, Alec and Kim, Jong Wook and Xu, Tao and Brockman, Greg and McLeavey, Christine and Sutskever, Ilya},
  booktitle=icml,
  year={2023}
}

@inproceedings{rafailov2023direct,
  title={Direct preference optimization: Your language model is secretly a reward model},
  author={Rafailov, Rafael and Sharma, Archit and Mitchell, Eric and Manning, Christopher D and Ermon, Stefano and Finn, Chelsea},
  booktitle=neurips,
  year={2023}
}

@inproceedings{rei2022cometkiwi,
  title={{CometKiwi}: {IST-Unbabel} 2022 submission for the quality estimation shared task},
  author={Rei, Ricardo and Treviso, Marcos and Guerreiro, Nuno M and Zerva, Chrysoula and Farinha, Ana C and Maroti, Christine and De Souza, Jos{\'e} GC and Glushkova, Taisiya and Alves, Duarte and Coheur, Luisa and others},
  booktitle=wmt,
  year={2022}
}

@inproceedings{saboo2019integration,
  title={Integration of dubbing constraints into machine translation},
  author={Saboo, Ashutosh and Baumann, Timo},
  booktitle=wmt,
  year={2019}
}

@inproceedings{saeki2022utmos,
  title={{UTMOS}: {UTokyo-SaruLab} system for {VoiceMOS} challenge 2022},
  author={Saeki, Takaaki and Xin, Detai and Nakata, Wataru and Koriyama, Tomoki and Takamichi, Shinnosuke and Saruwatari, Hiroshi},
  booktitle=is,
  year={2022}
}

@inproceedings{sahipjohn2024dubwise,
  title={{DubWise}: Video-guided speech duration control in multimodal {LLM}-based text-to-speech for dubbing},
  author={Sahipjohn, Neha and Gudmalwar, Ashishkumar and Shah, Nirmesh and Wasnik, Pankaj and Shah, Rajiv Ratn},
  booktitle=is,
  year={2024}
}

@inproceedings{salesky2021multilingual,
  title={The multilingual {TEDx} corpus for speech recognition and translation},
  author={Salesky, Elizabeth and Wiesner, Matthew and Bremerman, Jacob and Cattoni, Roldano and Negri, Matteo and Turchi, Marco and Oard, Douglas W and Post, Matt},
  booktitle=is,
  year={2021}
}

@inproceedings{saragih2025beyond,
  title={Beyond One-Shot Dubbing: Leveraging N-Best Translation and Prompted Paraphrasing with Synchrony-Aware Re-Ranking},
  author={Saragih, Jan Meyer and Mehmood, Faisal and Sakti, Sakriani},
  booktitle=apsipa,
  year={2025}
}

@article{seamless2025joint,
  title={Joint speech and text machine translation for up to 100 languages},
  author={{Seamless Communication Team}},
  journal={Nature},
  year={2025}
}

@misc{silerovad,
  title={Silero {VAD}: pre-trained enterprise-grade Voice Activity Detector ({VAD}), Number Detector and Language Classifier},
  author={{Silero Team}},
  year={2024},
  howpublished={\url{https://github.com/snakers4/silero-vad}},
}

@article{sodagar2011mpeg,
  title={The {MPEG-DASH} standard for multimedia streaming over the internet},
  author={Sodagar, Iraj},
  journal={IEEE MultiMedia},
  year={2011}
}

@article{spiteri2021english,
  title={English-language dubbing: challenges and quality standards of an emerging localisation trend},
  author={Spiteri Miggiani, Giselle},
  journal={The Journal of Specialised Translation},
  year={2021}
}

@article{spiteri2023quality,
  title={Quality in translation and adaptation for dubbing: Applied research in a professional setting},
  author={Spiteri Miggiani, Giselle},
  journal={The Journal of Specialised Translation},
  year={2023}
}

@inproceedings{sung2025voicecraft,
  title={{VoiceCraft-Dub}: Automated video dubbing with neural codec language models},
  author={Sung-Bin, Kim and Choi, Jeongsoo and Peng, Puyuan and Chung, Joon Son and Oh, Tae-Hyun and Harwath, David},
  booktitle=iccv,
  year={2025}
}

@inproceedings{tam2022isochrony,
  title={Isochrony-aware neural machine translation for automatic dubbing},
  author={Tam, Derek and Lakew, Surafel M and Virkar, Yogesh and Mathur, Prashant and Federico, Marcello},
  booktitle=is,
  year={2022}
}

@article{wang2025syncvoice,
  title={{SyncVoice}: Towards Video Dubbing with Vision-Augmented Pretrained {TTS} Model},
  author={Wang, Kaidi and He, Yi and Guan, Wenhao and Wu, Weijie and Ding, Hongwu and Zhang, Xiong and Wu, Di and Meng, Meng and Luan, Jian and Li, Lin and others},
  journal={arXiv preprint arXiv:2512.05126v1},
  year={2025}
}

@inproceedings{wu2023videodubber,
  title={{VideoDubber}: Machine translation with speech-aware length control for video dubbing},
  author={Wu, Yihan and Guo, Junliang and Tan, Xu and Zhang, Chen and Li, Bohan and Song, Ruihua and He, Lei and Zhao, Sheng and Menezes, Arul and Bian, Jiang},
  booktitle=aaai,
  year={2023}
}

@article{xu2025qwen25,
  title={{Qwen2.5-Omni} Technical Report},
  author={Xu, Jin and Guo, Zhifang and He, Jinzheng and Hu, Hangrui and He, Ting and Bai, Shuai and Chen, Keqin and Wang, Jialin and Fan, Yang and Dang, Kai and others},
  journal={arXiv preprint arXiv:2503.20215},
  year={2025}
}

@article{xu2026x,
  title={{X-Voice}: Enabling Everyone to Speak 30 Languages via Zero-Shot Cross-Lingual Voice Cloning},
  author={Xu, Rixi and Liu, Qingyu and Li, Haitao and Chen, Yushen and Niu, Zhikang and Yang, Yunting and Zhao, Jian and Li, Ke and Sisman, Berrak and Cheng, Qinyuan and others},
  journal={arXiv preprint arXiv:2605.05611},
  year={2026}
}

@article{yang2025infinitetalk,
  title={{InfiniteTalk}: Audio-driven video generation for sparse-frame video dubbing},
  author={Yang, Shaoshu and Kong, Zhe and Gao, Feng and Cheng, Meng and Liu, Xiangyu and Zhang, Yong and Kang, Zhuoliang and Luo, Wenhan and Cai, Xunliang and He, Ran and others},
  journal={arXiv preprint arXiv:2508.14033},
  year={2025}
}

\end{document}